\documentclass[letterpaper, 10pt, conference]{ieeeconf}  
\IEEEoverridecommandlockouts                              
\usepackage{duckuments}

\usepackage{algorithm}
\usepackage{balance}

\usepackage{graphicx}
\usepackage{epstopdf}

\usepackage{colortbl}

\usepackage{subcaption}

\usepackage{todonotes}

\usepackage{booktabs}

\usepackage{booktabs}
\usepackage{amsmath,amssymb}
\usepackage{graphicx}
\usepackage{dsfont}

\usepackage{amsmath,amssymb,amsfonts}
\usepackage{xcolor}
\usepackage{bm}
\usepackage{multirow}

\usepackage{cite}

\usepackage{amsmath}

\usepackage{scalerel}

\def\vecsign#1{\rule[1.388\LMex]{\dimexpr#1-2.5pt}{.36\LMpt}%
  \kern-6.0\LMpt\mathchar"017E}

\usepackage{stackengine,amsmath}
\stackMath

\title{\LARGE \bf
FORGE: Feasible Online Reduced-Order Reference Generation and Whole-Body Execution for Agile Humanoid Locomotion}

\author{Adrian~B.~Ghansah$^{1}$, Yujin~An$^{2}$, Jeeseop~Kim$^{3}$, Soon-Jo~Chung$^{1}$
\thanks{$^{1}$A. B. Ghansah, and S. J. Chung are with the Department of Control and Dynamical Systems, California Institute of Technology, Pasadena, CA 91125, USA, {\tt\small \{aghansah, sjchung\}@caltech.edu}\newline
$^{2}$Y. An is with the Department of Bioengineering, California Institute of Technology, Pasadena, CA 91125, USA, {\tt\small \{yan2\}@caltech.edu}\newline
$^{3}$J. Kim is with the Dept. of Aerospace and Mechanical Engineering, The University of Texas at El Paso, TX 79968, USA, {\tt\small jkim16@utep.edu}
}%
}

\begin{document}

\maketitle
\thispagestyle{empty}
\pagestyle{empty}

\begin{abstract}
    This paper presents a two-stage learning framework for versatile and agile humanoid locomotion that couples online reduced-order model (ROM) reference generation with whole-body tracking control. In the first stage, we train a command-conditioned whole-body reference generator using single rigid body dynamics combined with full humanoid kinematics. In the second stage, the frozen ROM policy is rolled out online to generate references for training a whole-body tracking policy on the full-order dynamics model. Learning a kino-dynamically consistent reference generator in the simplified ROM space enables us to generate references over a broad command space and over multiple locomotion strategies using a single unified framework. 
    We demonstrate the approach on a Unitree G1 humanoid robot, achieving longitudinal velocities from $-2.0~\mathrm{m/s}$ to $2.8~\mathrm{m/s}$, lateral velocities up to $\pm 1.0~\mathrm{m/s}$, angular velocities up to $\pm 3.0~\mathrm{rad/s}$, one-legged hopping, and continuous transitions across the commands, all using one unified reference policy and one whole-body tracking policy.
\end{abstract}

\section{Introduction}
\label{sec:introduction}

Recent advances have enabled humanoid robots to perform individual dynamic behaviors with impressive reliability. However, unifying these behaviors within a single control framework that is both agile and versatile remains challenging due to the high-dimensional, underactuated, and nonlinear hybrid dynamics of humanoids.

Model-based controllers often address this challenge by decomposing locomotion into two subproblems: generating reference trajectories and whole-body tracking. For reference generation, full-order model (FOM) trajectory optimization employs the complete dynamics, yielding trajectories that are dynamically consistent with the full-order system. However, using the complete dynamics may make the optimization problem computationally expensive \cite{reher2016realizing, ma2017bipedal, ramezani2014performance}.
Many frameworks therefore perform trajectory optimization offline or update the planner at a lower rate than the tracking controller \cite{reher2016realizing, esteban2025reduced}, which may limit their applicability in online settings. 

Reduced-order models (ROMs) offer an alternative approach in which the robot is represented by a simplified model capturing its dominant dynamics. 
Because this low-dimensional representation involves fewer states and constraints, the associated optimization is less expensive, enabling ROM-based planners to generate references online, making them attractive for real-time planning \cite{xiong2021global, xiong20223}. 
However, a trajectory that is feasible for the ROM may not be feasible for the full-order system, as the low-dimensional representation does not account for full-body kinematic constraints and limb dynamics \cite{ghansah2025hierarchical, xiong20223}.

\begin{figure}
    \centering
    \includegraphics[width=1.0\linewidth]{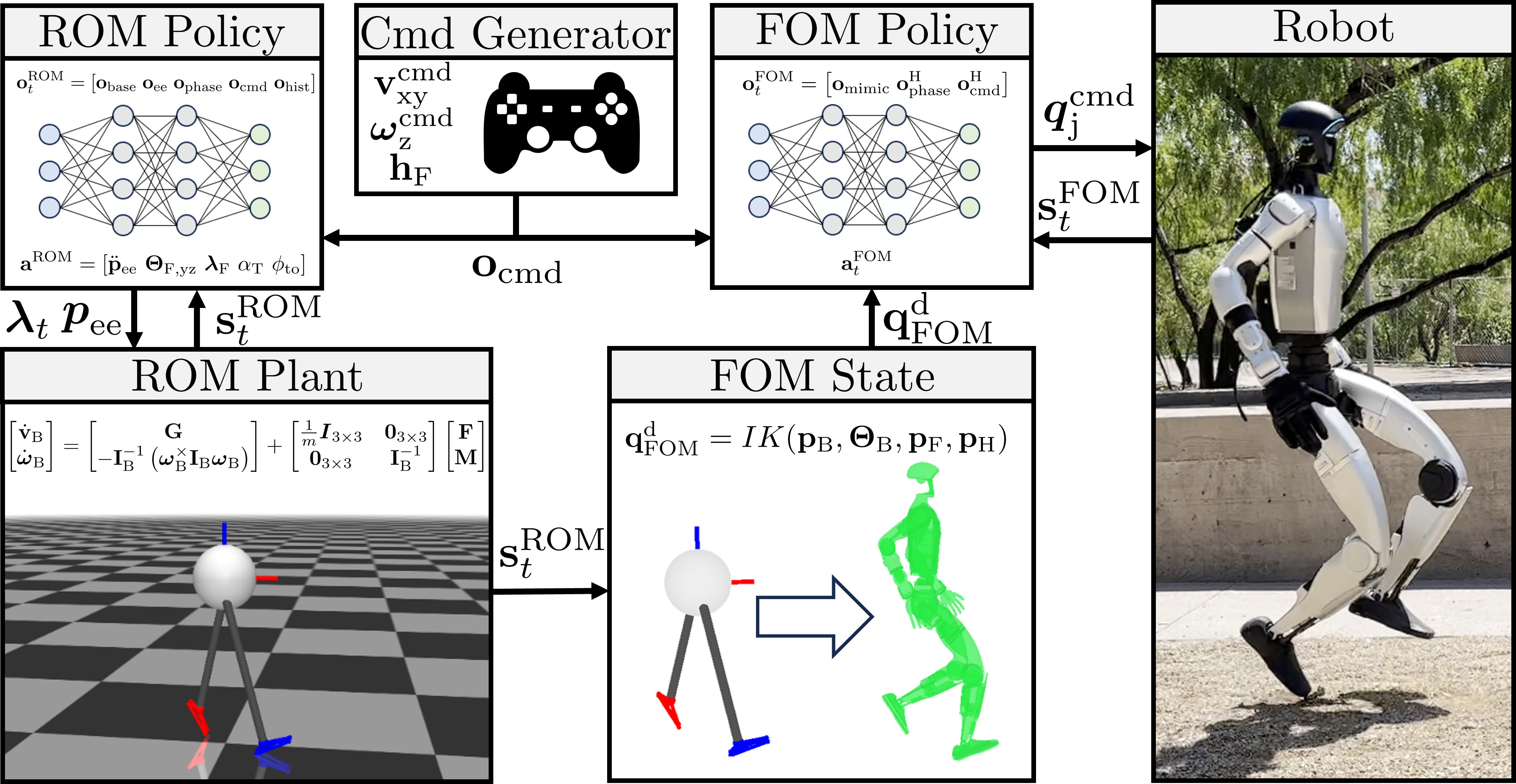}
    \caption{A command generator drives the ROM controller and plant rollouts, from which we generate whole-body references that are passed to a whole-body tracking controller.}
    \label{fig:hero}
    \vspace{-0.15cm}
\end{figure}

Learning-based controllers provide an alternative approach for handling the complex dynamics of humanoid locomotion. Rather than relying on an explicit trajectory optimizer or a ROM planner to compute control actions, reinforcement learning (RL) trains policies that map observations directly to actions by maximizing a reward function. Massively parallel simulation allows these policies to accumulate large amounts of experience through training, and end-to-end policies trained in this manner have demonstrated remarkable robustness and task performance \cite{radosavovic2024real, siekmann2021sim, van2024revisiting}. 
Locomotion frameworks that learn distinct skills via separately trained policies have been explored \cite{li2025reinforcement}, yet, training a single policy that covers multiple tasks or broad command ranges remains difficult.

Beyond task rewards alone, a complementary strategy is to supply reference trajectories during training, providing structured objectives for RL training \cite{liao2025beyondmimic, sleiman2026zest}. Whole-body tracking policies trained in this way have demonstrated highly agile behaviors \cite{he2025asap, wang2026spark, sleiman2026zest}. 
However, this agility depends on high-quality reference trajectories \cite{wang2026spark, he2025asap, sleiman2026zest}, which remain challenging to obtain over a broad command space, let alone generate online \cite{liao2025beyondmimic, sleiman2026zest}.

This work addresses the challenge of generating and tracking high-quality whole-body reference trajectories online for versatile and agile humanoid locomotion over broad command spaces. To this end, we propose a two-stage framework. In the first stage, we train a command-conditioned whole-body reference generator based on single rigid body (SRB) dynamics combined with full-order humanoid kinematics. In the second stage, we integrate this reference generator into the training of a whole-body tracking policy.

We show that the proposed framework generates high-quality whole-body reference trajectories over a broad command space and leverages them to train a whole-body tracking policy. We further validate the framework on a Unitree G1 humanoid robot.
Our contributions are:
\begin{itemize}
    \item A command-conditioned whole-body reference generator that combine SRB dynamics with full-order humanoid kinematics to produce dynamically informed references over a broad command space.
    \item A two-stage training framework that integrates the online reference generator into the training of a whole-body tracking policy.
    \item A hardware demonstration of the proposed framework, achieving longitudinal velocities from $-2.0~\mathrm{m/s}$ to $2.8~\mathrm{m/s}$, lateral velocities up to $\pm 1.0~\mathrm{m/s}$, angular velocities up to $\pm 3.0~\mathrm{rad/s}$, one-legged hopping, and continuous transitions across the commands.
\end{itemize}

\section{Related Work}
\label{sec:related_work}

\subsection{Reduced-Order Models for Humanoid Locomotion}


ROMs are attractive for humanoid locomotion control because they provide simplified representations of the complex full-order humanoid dynamics. These models are commonly employed to plan foothold locations and to generate the ground reaction forces and moments (GRFMs) required to stabilize the robot via real-time control. 
One of the canonical ROMs is the Linear Inverted Pendulum (LIP) model \cite{kajita2003biped, kajita20013d}, which represents the robot as a point-mass moving in the horizontal plane at a fixed center of mass (COM) height, rotating about a passive pivot point at the foot. 
Related ROMs capture additional features of different locomotion behaviors: the Hybrid Linear Inverted Pendulum (HLIP) model \cite{xiong20223} incorporates the hybrid nature of legged locomotion, the Angular Linear Inverted Pendulum (ALIP) model \cite{gong2020angular, dosunmu2023stair} accounts for angular momentum about the stance foot, and the Spring-Loaded Inverted Pendulum (SLIP) model \cite{blickhan1989spring} permits variation in vertical velocity. 
These pendulum models, however, are primarily concerned with the regulation of step length and step timing; centroidal models instead incorporate both linear and angular momentum dynamics, providing a richer representation of the robot system. One such model is the SRB model \cite{di2018dynamic}, which is commonly coupled with Raibert-style footstep planners \cite{raibert1986legged}, and to which model predictive control (MPC) is often applied to compute GRFMs that are then mapped to feedforward torques through the contact Jacobian \cite{di2018dynamic, li2021force, ghansah2025hierarchical}. 

ROMs have also been employed to guide the training of RL policies, often through objectives derived from ROM outputs, such as step-length targets. In \cite{lee2024integrating}, the LIP model was used to generate rewards based on desired footstep targets, whereas in \cite{dai2026walk, kim2025learning}, the ALIP model was used to generate step-length rewards that regulate the angular momentum for traversing constrained footholds. The HLIP model was employed in \cite{liu2026hlip} to generate step-length rewards for achieving multigait locomotion and high-speed running. Common to these works is that the ROMs guide the policy during training rather than generating whole-body reference trajectories for the robot to track. Offline-generated ROM trajectories have also been explored for policy training. In \cite{green2021learning}, gait libraries based on the SLIP model were generated offline via trajectory optimization, and the resulting references were used during training to provide COM velocity and foot location rewards. Similarly, in \cite{batke2022optimizing}, trajectory libraries based on the SRB model were generated offline via optimization for various dynamic maneuvers and were subsequently used as expert references during policy training to generate COM linear velocity, angular momentum, and foot position rewards.
The direction most closely related to the proposed work is \cite{compton2026terrain}, in which LIP-based whole-body references are generated online. However, whereas \cite{compton2026terrain} focuses on whole-body reference generation for various terrain setups, 
our work proposes a two-stage framework that broadens the command space, enabling whole-body locomotion that ranges from agile maneuvers to high-speed running while also accommodating seamless transitions between commands.

\subsection{Humanoid Reference Generation}

In humanoid locomotion, providing motion references, such as kinematic and joint-torque trajectories, as a prior for imitation-based RL has proven crucial for robust policy learning of online tracking controllers. Model-based trajectory optimization serves as a foundation of such reference generation by formulating motion synthesis as a constrained optimization problem that enforces physically grounded dynamics to produce high-quality, dynamically feasible trajectories. These optimization-based approaches range from formulations built on ROMs~\cite{batke2022optimizing} to those built on full-order-dynamics~\cite{liu2024opt2skill}. However, classical optimization pipelines typically rely on precomputed gait libraries and separate per-task policies, which limits versatility and hinders transitions between behaviors. Human motion data offers an alternative source for the reference generation, yet retargeting motions from human to robot morphologies often yields kinematically or dynamically infeasible references. Consequently, obtaining high-quality references for diverse downstream tasks remains an important open problem~\cite{wang2026spark, sleiman2026zest,liao2025beyondmimic}. 

\begin{figure*}
    \centering
    \includegraphics[width=1.0\linewidth]{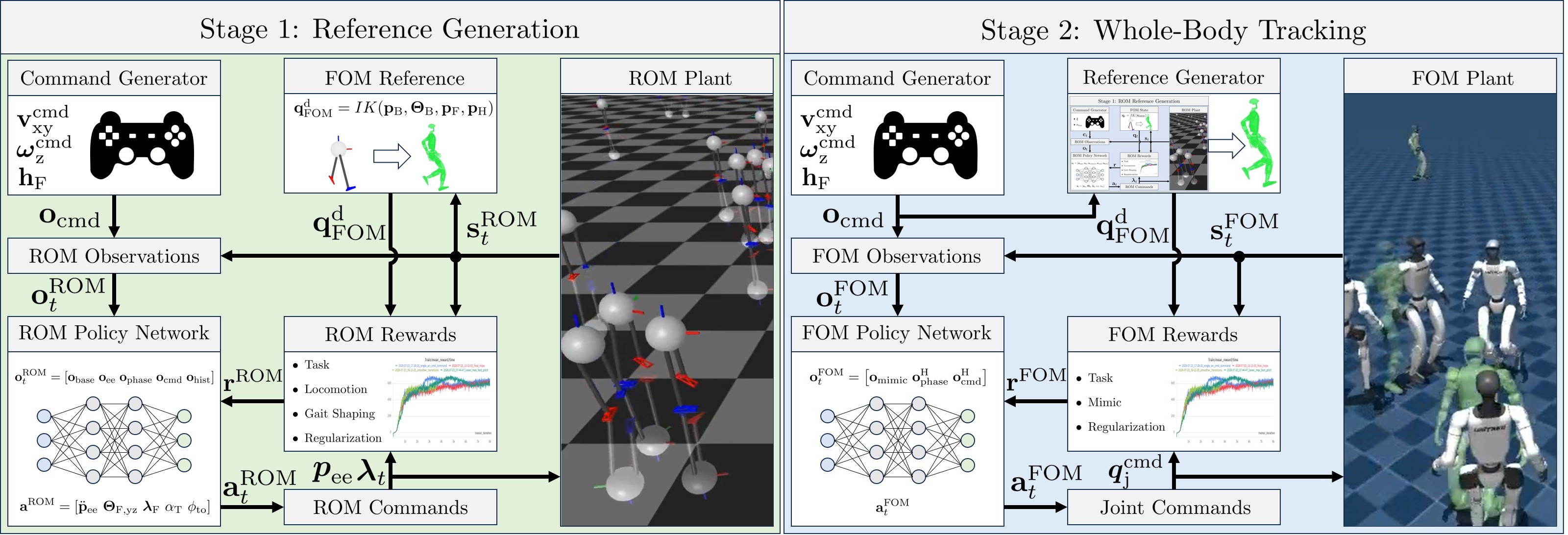}
    \caption{An overview of the two-stage training architecture. 
    (Left) Illustrates the Stage 1 training, in which a command-conditioned whole-body reference generator is trained based on SRB dynamics and the full humanoid kinematics within a custom ROM simulator.  
    (Right) Illustrates the Stage 2 training, in which a whole-body tracking policy for the full order system is trained to track the references generated in Stage 1.}
    \label{fig:training_architecture}
    \vspace{-1.5em}
\end{figure*}

\subsection{Whole-Body Motion Tracking}
Whole-body motion tracking for humanoids aims to transform kinematic references into dynamically feasible motions, enabling humanoids to operate in real environments through imitation-based RL. The control policy is trained to minimize the discrepancy between the reference motion and the executed motion, which allows the humanoid to reproduce diverse behaviors while satisfying physical constraints. Reference motions are typically drawn from large-scale human datasets. This paradigm was first demonstrated for humanoid character animation in physical simulation~\cite{peng2018deepmimic, chentanez2018physics}, and has since been extended to teach real humanoid robots skills ranging from various locomotion behaviors~\cite{liao2025beyondmimic} to whole-body scene interactions~\cite{yang2025omniretarget}. Real-time human imitation via teleoperation further broadens deployment to a wide range of environments~\cite{ze2025twist, ze2025twist2, he2024h2o}. To move beyond skills confined to fixed environments and toward real-time control in dynamic settings, a generator--tracker hierarchy has recently been proposed~\cite{zhang2026wholebody, xu2025parc, serifi2024robotmdm}, where a diffusion-based motion generator produces future whole-body trajectories conditioned on task inputs and a tracker converts them into physically executable actions. However, all of these paradigms fundamentally rely on large-scale human motion datasets, which remain scarce and expensive to acquire. 

\section{Background}

\subsection{Single Rigid Body Dynamics}

The SRB model can effectively capture both the linear and rotational dynamics of a floating base system by ignoring the limbs and treating the whole system as a lumped mass with fixed inertia. Using Newton-Euler dynamics, we can write the SRB dynamics as:
\begin{equation}
    \label{eq:newton_euler}
    \begin{bmatrix}
        \dot{\mathbf{v}}_\textnormal{B} \\
        \dot{\boldsymbol{\omega}}_\textnormal{B}
    \end{bmatrix}
    =
    \begin{bmatrix}
        \mathbf{G}\\
        -\textbf{I}_\textnormal{B}^{-1} \left( \boldsymbol{\omega}_\textnormal{B}^\times \textbf{I}_\textnormal{B} \boldsymbol{\omega}_\textnormal{B} \right)
    \end{bmatrix}
    +
    \begin{bmatrix}
        \frac{1}{m} \boldsymbol{I}_{3 \times 3} & \mathbf{0}_{3 \times 3} \\
        \mathbf{0}_{3 \times 3} & \textbf{I}_\textnormal{B}^{-1} 
    \end{bmatrix}
    \begin{bmatrix}
        \mathbf{F} \\
        \mathbf{M}
    \end{bmatrix},
\end{equation}
where ${\mathbf{v}}_\textnormal{B} \in \mathbb{R}^3$ and ${\boldsymbol{\omega}}_\textnormal{B} \in \mathbb{R}^3$ represent the linear and angular velocity of the body, $\mathbf{G} = [0 \ 0 \ -g]^\top$ with gravitational acceleration $g$, and $\mathbf{F} \in \mathbb{R}^3$ and $\mathbf{M} \in \mathbb{R}^3$ represent the sum of external forces and moments acting on the system. Finally, we have that $m \in \mathbb{R}$ and $\mathbf{I}_\textnormal{B} \in \mathbb{R}^{3 \times 3}$ are the mass and moment of inertia around COM. The full SRB state can therefore be expressed as:
\begin{equation}
    \label{eq:srb_state}
    \mathbf{x}_\textnormal{SRB} 
    = 
    \begin{bmatrix}
    \textbf{p}_\textnormal{B} & \boldsymbol{\Theta}_\textnormal{B} & \textbf{v}_\textnormal{B} & \boldsymbol{\omega}_\textnormal{B}
    \end{bmatrix},
\end{equation}
where $\textbf{p}_\textnormal{B} \in \mathbb{R}^3$ and $\boldsymbol{\Theta}_\textnormal{B} \in \mathbf{SO}(3)$ represent the body's position and orientation relative to the inertial frame. 

Let the GRFMs for the left and right feet be given by:
\begin{equation}
    \boldsymbol{\lambda}_\textnormal{F} 
    = 
    \left[ \mathbf{F}_\textnormal{LF}^\top \ \mathbf{F}_\textnormal{RF}^\top \ \mathbf{M}_\textnormal{LF}^\top \ \mathbf{M}_\textnormal{RF}^\top \right]^\top
    \in
    \mathbb{R}^{12}.
\end{equation}
Then the external forces in \eqref{eq:newton_euler} are given by: $\mathbf{F}=\sum_i \mathbf{F}_i$ and $\mathbf{M}=\sum_i(\mathbf{r}_i^\times \mathbf{F}_i + \mathbf{M}_i)$,
where $i\in\{\textrm{LF}, \textrm{RF}\}$ and $\mathbf{r}_\textnormal{LF}$ and $\mathbf{r}_\textnormal{RF}$ is the vector from the base to left and right foot.


\subsection{Reinforcement Learning}
RL formulates control as the search for an optimal policy
$\pi^*$ that maximizes the expected discounted return $\mathbb{E}_{\mathcal{T} \sim p_{\pi}(\mathcal{T})}
    \left[ \sum_{t=0}^{N_{\textnormal{ep}}} \gamma^{t} r(\mathbf{s}_{t}, \mathbf{a}_{t}) \right],
    \label{paper05:eq:return}$
where $\mathbf{s}_{t}$ is the full environment state, $\mathbf{a}_{t}$ the agent action, and
$r$ the reward at each time step $t$. The episode terminates at step $N_{\textnormal{ep}}$,
$\gamma \in [0,1)$ discounts future rewards, 
and $\mathcal{T}$ denotes a
trajectory induced by $\pi$. 
The problem is modeled as a Partially Observable Markov Decision Process
(POMDP), in which $\mathbf{s}_{t}$ governing the dynamics is not  fully observable from humanoid onboard sensing. The policy instead acts on an
observation $\mathbf{o}_{t} = \boldsymbol{\Omega}(\mathbf{s}_{t})$, where $\boldsymbol{\Omega}$ is the observation
function producing a partial and noisy projection of $\mathbf{s}_{t}$ from
proprioception. Since a single
observation often is insufficient to disambiguate $\mathbf{s}_{t}$, the policy is normally
conditioned on a history of length $h$ comprising past observations and
actions, together with a user control command $\mathbf{o}_{\textnormal{cmd}}$, giving
$ \pi\left( \mathbf{a}_{t} \mid \mathbf{o}_{t-h:t},\, \mathbf{a}_{t-h-1:t-1},\,
    \mathbf{o}_{\textnormal{cmd}}\right).
    \label{paper05:eq:policy} $

\section{Whole-Body Reference Generation}

\subsection{Overview}

The objective of the Stage 1 illustrated in Fig. \ref{fig:training_architecture}, is to develop a whole-body reference generator capable of realizing a broad range of locomotion commands through a single policy. Learning multiple skills within one policy while ensuring smooth transitioning remains challenging. We therefore employ the SRB dynamics in \eqref{eq:newton_euler} and incorporate the full humanoid kinematics information into the training process, ensuring that all generated motions respect the kinematic capabilities of the robot. 
The Stage 1 training is conducted in a numerical ROM simulator, as described in Section \ref{sec:stage_1_training_environment}.
The output of the Stage 1 is a command-conditioned whole-body reference generator, which is subsequently employed by the Stage 2 whole-body tracker.


\subsection{Training Environment}
\label{sec:stage_1_training_environment}

Stage 1 training is conducted in a proposed numerical ROM simulation environment that combines SRB dynamics with full humanoid kinematics using simplified contact kinematics. Because the environment is formulated directly at the reduced-order level, it exposes the quantities that ROM-based control acts upon, specifically the end-effector states governed by task-space accelerations and contact timings as well as applied GRFMs. This formulation therefore provides immediate access to the command inputs without mediation by an external physics or contact engine. The environment instead advances the system states through numerical evaluation that encompasses both contact and collision detection, with MuJoCo \cite{todorov2012mujoco} employed solely for visualization.

\subsubsection{Kinematics and Dynamics}
In our simulator, only the robot base is modeled dynamically, while the end effectors are represented kinematically. The inertial properties of the physical robot are lumped into a single rigid body and the simulated physics evolves according to the SRB dynamics in \eqref{eq:newton_euler}. The external forces and moments acting on the system are computed such that the GRFMs are included only when the corresponding foot is in contact with the ground. The end effectors of the ROM comprise the feet poses and twists together with the hand positions and velocities, and they are driven kinematically through end effector task space accelerations. Given the ROM end effector poses and the SRB pose, the generalized coordinates of the full robot, $\mathbf{q}_\textnormal{FOM} = [\mathbf{p}_\textnormal{B} \ \boldsymbol{\Theta}_\textnormal{B} \ \mathbf{q}_\textnormal{j}]$, are obtained via inverse kinematics, where $\mathbf{q}_\textnormal{j} \in \mathbb{R}^{n_j}$ are the joint positions of the robot and $n_j$ denotes the number of joints. The joint velocities and accelerations are subsequently computed through numerical differentiation of the joint positions. Although the ROM assumes massless limbs, the leg joint torques, $\boldsymbol{\tau}_\textnormal{legs}$, are approximated using the contact Jacobian, $\mathbf{J}_\textnormal{c}$, together with the applied GRFMs as $\boldsymbol{\tau}_\textnormal{legs} = \mathbf{J}_\textnormal{c}^\top \boldsymbol{\lambda}_\textnormal{F}$.

\subsubsection{Contact Kinematics}
\label{sec:contact_model}
Each foot is modeled as an ankle with four contact points, two at the front and two at the back of the foot, matching the geometry of the robot's foot. When a foot contact point is active, it is fixed in space, while the foot remains free to rotate about that point. We choose to constrain the foot roll to zero rotation to avoid brittle side contacts. As a result, the two heel contact points are always in the same contact state, and the same for the two toe contact states. Consequently, a foot in contact cannot yaw, however it is free to pitch around the toes and heels. Figure \ref{fig:gait_phasing}(a) shows the possible contact states for a foot in our simulator.

\subsubsection{Gait Phasing}
\begin{figure}
    \includegraphics[width=0.48\textwidth]{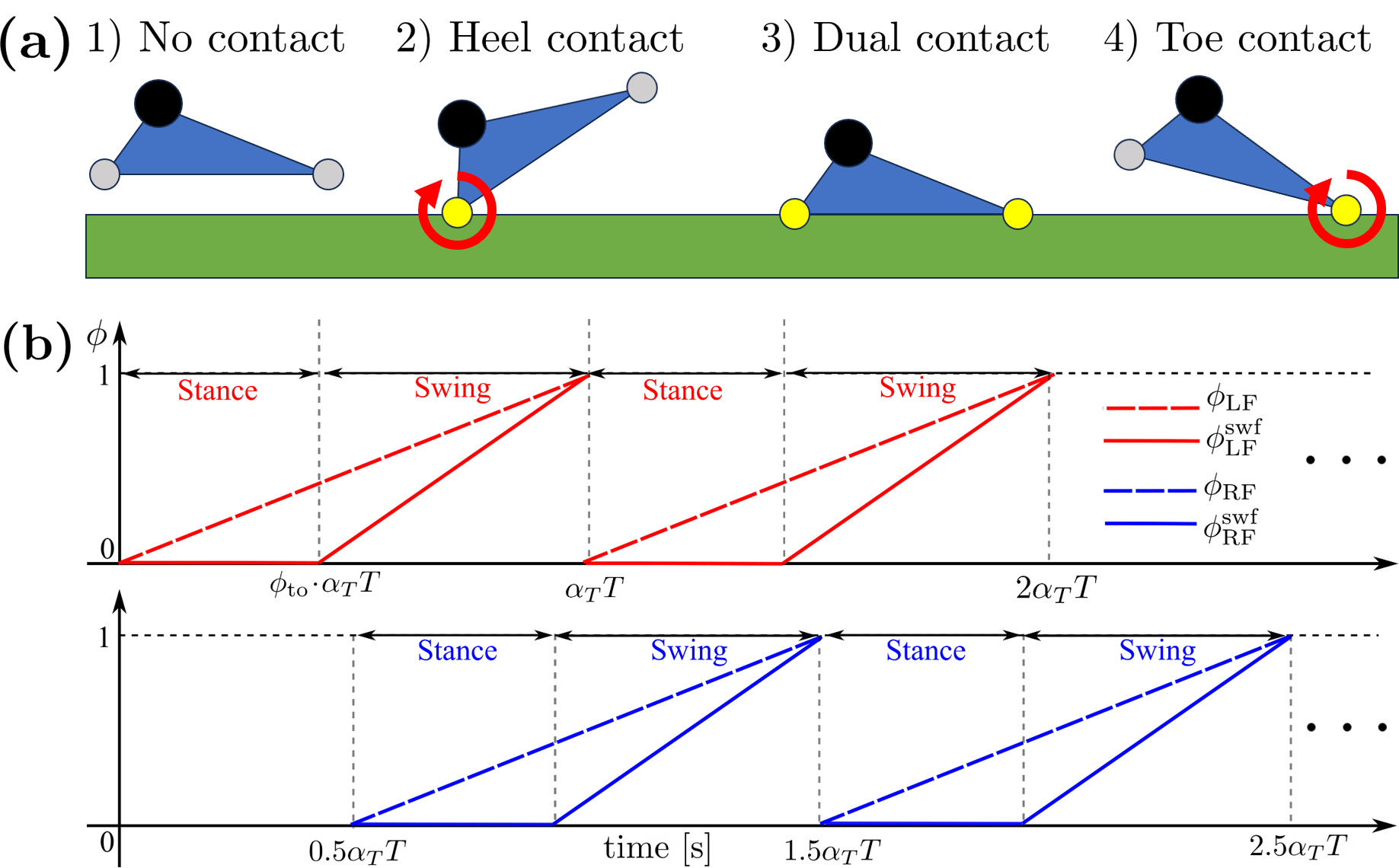}
    \caption{(a) The different ROM foot contact modes.
    (b) The relationship between the swing foot phase and gait phase.}
    \label{fig:gait_phasing}
\end{figure}
Foot contacts are gated via a phasing variable $\phi \in [0, 1)$ which increases monotonically from 0 to 1 over a full gait cycle before resetting to 0 (see Fig. \ref{fig:gait_phasing}(b)). For the left foot (LF), the phasing variable evolves as
\begin{equation}
    \label{eq:gait_phasing_lf}
    \phi_{\textnormal{LF},t+\Delta t}
    =
    \operatorname{mod}\!\left(\phi_{\textnormal{LF},t} + \tfrac{\Delta t}{\alpha_T T},\, 1\right),
\end{equation}
where $T \in \mathbb{R}^+$ is the nominal gait period over two steps and $\alpha_T \in \mathbb{R}^+$ is a scaling factor of the step period that allows us to slow down or speed up the gait. To ensure that the left and right feet remain exactly out of phase, the phasing variable of the right foot (RF) is defined as a function of that of the left foot with a phase offset of 0.5, that is,
\begin{equation}
    \label{eq:gait_phasing_rf}
    \phi_{\textnormal{RF},t+\Delta t}
    =
    \operatorname{mod}\!\left(\phi_{\textnormal{LF},t+\Delta t} + 0.5,\, 1\right).
\end{equation}
When the phasing variable exceeds a take-off threshold $\phi_\textnormal{to} \in (0, 1)$, the corresponding foot is considered to be in the swing phase. Otherwise, the foot is considered to be in the stance phase, meaning that it is in contact with the ground. The swing foot phasing variables $\phi_\textnormal{LF}^{\textnormal{swf}}, \phi_\textnormal{RF}^{\textnormal{swf}} \in [0, 1)$ denote the progress of the swing feet and are given by:
\begin{equation}
    \label{eq:gait_phasing_swf}
    \phi_{\textnormal{LF/RF}}^{\textnormal{swf}}
    =
    \begin{cases}
        0, & \phi_{\textnormal{LF/RF}}  < \phi_\textnormal{to} \\[4pt]
        \dfrac{\phi_{\textnormal{LF/RF}} - \phi_\textnormal{to}}{1 - \phi_\textnormal{to}}, & \phi_{\textnormal{LF/RF}}  \ge \phi_\textnormal{to}
    \end{cases}
\end{equation}
Figure \ref{fig:gait_phasing}(b) illustrates the gait phasing variables and swing foot phasing variables.

\subsection{ROM MDP Formulation}

We now present the MDP formulation of the ROM training. Specifically, we describe the states of the environment, the observations, 
the action space,
and the reward structure.

\subsubsection{State}

The full state of the environment is defined as a combination of the SRB states, the end effector poses and twists, and the phasing variables:
\begin{equation}
    \label{eq:rom_training_state}
    \mathbf{s}^\textnormal{ROM} = [\mathbf{x}_\textnormal{SRB} \ 
    \mathbf{p}_\textnormal{F} \ 
    \mathbf{p}_\textnormal{H} \ 
    \boldsymbol{\Theta}_\textnormal{F} \ 
    \mathbf{v}_\textnormal{F} \ 
    \mathbf{v}_\textnormal{H} \ 
    \boldsymbol{\omega}_\textnormal{F} \ 
    \boldsymbol{\Phi}],
\end{equation}
where $\mathbf{p}_\textnormal{F} \in \mathbb{R}^6$ and $\mathbf{p}_\textnormal{H} \in \mathbb{R}^6$ denote the two foot positions and hand positions in the world frame, and $\mathbf{v}_\textnormal{F} \in \mathbb{R}^6$ and $\mathbf{v}_\textnormal{H} \in \mathbb{R}^6$ denote the linear velocities. Furthermore, we have that  $\boldsymbol{\Theta}_\textnormal{F,yz} = \left[\theta_\textnormal{LF} \ \theta_\textnormal{RF} \ \psi_\textnormal{LF} \ \psi_\textnormal{RF} \right] \in \mathbb{R}^4$ denotes the left and right foot pitch and yaw in the world frame, and $\boldsymbol{\omega}_\textnormal{F,yz} \in \mathbb{R}^4$ the feet angular y and z velocities.
The vector $\boldsymbol{\Phi} = [\phi_\textnormal{LF} \ \phi_\textnormal{RF} \ \phi_\textnormal{LF}^\textnormal{swf} \ \phi_\textnormal{RF}^\textnormal{swf}] \in \mathbb{R}^4$ collects the phasing variables of the left and right feet with the corresponding swing foot phasing variables.

\subsubsection{Observation}

The ROM policies receives an observation consisting of the following groups:
\begin{equation}
    \label{eq:rom_observation}
    \begin{aligned}
        \mathbf{o}_t^\textnormal{ROM}  = 
        [
            \mathbf{o}_{\mathrm{base}} \;
            \mathbf{o}_{\mathrm{ee}} \;
            \mathbf{o}_{\mathrm{phase}} \;
            \mathbf{o}_{\mathrm{cmd}} \;
            \mathbf{o}_{\mathrm{hist}}
        ].
    \end{aligned}
\end{equation}

The base observation is given by $\mathbf{o}_{\mathrm{base}} = \left[\mathbf{x}_\textnormal{upr} \ \mathbf{v}_\textnormal{B} \ \boldsymbol{\omega}_\textnormal{B} \right] \in \mathbb{R}^9$, where we define the upright base pose states as $\mathbf{x}_{\mathrm{upr}}=\left[\Delta p_\textnormal{B,z} \ \varphi_\textnormal{B} \ \theta_\textnormal{B} \right] \in \mathbb{R}^3$. Here $\Delta p_\textnormal{B,z}$ is the base height relative to a nominal desired height. The base pose states enable the robot to track a upright posture, while the base twist states facilitate the tracking of twist commands and help mitigate velocity oscillations.

To enable the direct control of the end effector poses and twists via task space acceleration, the end effector observations $\mathbf{o}_{\mathrm{ee}} = \left[\mathbf{p}_\textnormal{F} \ \mathbf{p}_\textnormal{H} \ \mathbf{v}_\textnormal{F} \ \mathbf{v}_\textnormal{H} \ \boldsymbol{\Theta}_\textnormal{F,yz} \ \boldsymbol{\omega}_\textnormal{F,yz} \right] \in \mathbb{R}^{32}$ include the positions and linear velocities of the feet and hands relative to the base, expressed in the base control frame, together with the foot orientations in the base control frame and the foot angular velocities in the body frame. 
%
The phase observations $\mathbf{o}_{\mathrm{phase}} = \left[\mathbf{c}_\textnormal{F} \ \sin{(2\pi\boldsymbol{\Phi})} \ \cos{(2 \pi \boldsymbol{\Phi})} \right] \in \mathbb{R}^{10}$ complement the continuous motion information with the discrete foot-ground interaction, consisting of binary contact indicators for the left and right feet, $\mathbf{c}_\textnormal{F} \in \{0, 1\}^2$, and sines and cosines of the foot phasing variables, which locate the robot within the gait cycle. 
%
The desired behavior itself is specified by the command observations $\mathbf{o}_{\mathrm{cmd}} = \left[\mathbf{v}_{x,y}^\textnormal{cmd} \ \omega_\textnormal{z}^\textnormal{cmd} \ \mathbf{h}_\textnormal{F} \right] \in \mathbb{R}^7$, which contain the commanded base twist and hopping commands, $\mathbf{h}_\textnormal{F} \in \mathbb{R}^4$, two for each foot.
The twist command defines the velocity to be tracked, whereas two hopping commands per foot indicate when a foot should be in the air, with one applied immediately, and the other applied after a short delay preceding the deactivation of ground contact.
The history observation $\mathbf{o}_{\mathrm{hist}} = \left[ \mathbf{a}_{t-1} \ \mathbf{L}_\textnormal{prev} \right] \in \mathbb{R}^{32}$ comprise the previous action $\mathbf{a}_{t-1} \in \mathbb{R}^{28}$ and the most recently executed step lengths in the $x$ and $y$ directions, $\mathbf{L_\textnormal{prev}} \in \mathbb{R}^4$. This promotes smoother behavior with reduced action chattering.

\subsubsection{Action Space and Action Realization}

The actions are designed to regulate the ROM inputs associated with the end-effector states, the step timing, and the GRFMs. The action space of the ROM is defined as
\begin{equation}
    \label{eq:action_space}
    \mathbf{a}^\textnormal{ROM}
    = 
    \left[
    \ddot{\mathbf{p}}_\textnormal{ee} \ 
    \boldsymbol{\Theta}_\textnormal{F,yz} \ 
    \boldsymbol{\lambda}_\textnormal{F} \ 
    \alpha_\textnormal{T} \ 
    \phi_\textnormal{to}  
    \right]
    \in \mathbb{R}^{28},
\end{equation}
where, $\ddot{\mathbf{p}}_\textnormal{ee} \in \mathbb{R}^{10}$ contains the desired end-effector linear accelerations expressed in the body frame, except for the feet linear $z$ acceleration, which is instead governed by phase-based trajectories. 
$\boldsymbol{\Theta}_\textnormal{F,yz} \in \mathbb{R}^{4}$ contains the pitch and yaw angles of the feet which are commanded directly in the base control frame. Analogously, $\boldsymbol{\lambda}_\textnormal{F} \in \mathbb{R}^{12}$ contains the desired GRFMs for the left and right feet, expressed in the respective control frames of the feet. 
$\alpha_\textnormal{T}$ and ${\phi}_\textnormal{to}$ are gait phasing variables. The actions are first clipped to $\pm 1.0$ and subsequently mapped through linear interpolation onto the admissible ROM ranges. 

\begin{table}[t]
\centering
\caption{Reward terms used for ROM training.}
\label{tab:rom_rewards}
\footnotesize
\setlength{\tabcolsep}{2pt}
\begin{tabular}{@{}lcc@{}}
\toprule
\textbf{Reward Term} & \textbf{Expression} & \textbf{Weight} \\
\midrule
\textbf{Task} & & \\
Base lin vel tracking & $\exp\!(-\tfrac12\lVert \mathbf{v}_{\textnormal{xy}}-\mathbf{v}^\textnormal{cmd}_\textnormal{xy} \rVert^2/\sigma_{v_\textnormal{xy}}^2)$ & $16.0$ \\
Base ang vel tracking & $\exp\!(-\tfrac12\lVert \omega_\textnormal{z} - \omega_\textnormal{z}^\textnormal{cmd}\rVert^2/\sigma_{\omega_z}^2)$ & $8.0$ \\
\midrule
\textbf{Locomotion} & &  \\
Standing contact      & $\rho_{\mathrm{ds}}^2\,1_{\mathrm{st}}$ & $4.0$ \\
Step-period deviation & $((\alpha_T - 1)/0.4)^2$ & $-2.5$ \\
Contact-phase thresh. & $(\phi_\textnormal{to}-\phi_\textnormal{to}^\textnormal{d})^2$ & $-10.0$ \\
\midrule
\textbf{Gait Shaping} & & \\
Base pose ($i \in \mathbf{x}_{\mathrm{upr}}$) & $\exp\!(-\tfrac12(x_{\mathrm{i}}-x_{\mathrm{i}}^\textnormal{d})^2/\sigma_{\mathrm{i}}^2)$ & $5.0$ \\

Stance-foot flat      & $\bigl(\exp(-\|\boldsymbol{\Theta}_\textnormal{F,y}\|^2/(n_c\sigma_{\Theta_\textnormal{F,y}}^2))-1\bigr)1_{\mathrm{st}}$& $1.0$ \\

Hand pos. (BCF) & $\exp\!(-\tfrac12\lVert \mathbf{p}_\textnormal{H}-\mathbf{p}_\textnormal{H}^\textnormal{d}\rVert^2/\sigma_\textnormal{H}^2)$ & $10.0$ \\
\midrule
\textbf{Regularization} & & \\
Action-rate           & $\lVert \mathbf{a}_t^\textnormal{ROM} - \mathbf{a}_{t-1}^\textnormal{ROM}\rVert^2$ & $-1.0$ \\
Joint acceleration    & $\lVert\ddot {\mathbf{q}}\rVert^2 / n_j$ & $-3\times10^{-4}$ \\
Joint limit           & $\lVert(\underline{\mathbf{q}}-\mathbf{q}_t)^{+} + (\mathbf{q}_t-\bar{\mathbf{q}})^{+}\rVert^2$ & $-5.0$ \\
Self-collision        & $\#\{\text{penetrating capsule pairs}\}$ & $-2.0$ \\
End-effector accel.   & $\lVert\ddot{\mathbf{p}}_{\mathrm{ee}}\rVert^2$ & $-5\times10^{-3}$ \\
Base twist error & $\lVert\boldsymbol{\xi}-\boldsymbol{\xi}^{\mathrm{cmd}}\rVert^2$ & $-0.5$ \\
Standing motion & $\lVert \boldsymbol{\zeta} \rVert\,1_{\mathrm{st}}$ & $-1.0$ \\
Leg Torques      & $\lVert \bar{\boldsymbol{\tau}}_\textnormal{legs} \rVert^6$ & $-1.0$ \\
\bottomrule
\end{tabular}
\end{table}

\subsection{Reward Structure}

The set of rewards used to train the policy can broadly be divided into the following categories: 
task rewards, locomotion rewards, gait shaping rewards, and regularization rewards. 
The full overview is presented in Table ~\ref{tab:rom_rewards}.

The task rewards consists of the linear x/y tracking rewards and an angular velocity z tracking reward. 
These rewards use exponential kernels with $\sigma$ denoting the per-term Gaussian width in order to incentivize accurate tracking of the user commanded twist signals $(v_x^\textnormal{cmd}, v_y^\textnormal{cmd}, \omega_z^\textnormal{cmd})$.
As the main goal of the controller is to be able to track these commands, we give a relatively large weight to these terms. 

The locomotion rewards are used to control the step timing of the robot, 
that is when a foot should be in contact with the ground and when to be in the air. 
The standing contact reward encourages full ground support when the velocity command is near zero. $\rho_\textnormal{ds} \in \{0,  0.5, 1.0\}$ represents contact with zero feet, one foot, or both feet, respectively. 
$1_{\textnormal{st}} \in \{0, 1 \}$ is $1$ when the commanded velocity is close to zero and otherwise $0$.  
The step-period deviation term penalizes the policy for deviating from the nominal step period, driving the step-period scaling factor $\alpha_T$ toward 1. The contact phase threshold reward penalizes the policy for deviating from the nominal takeoff-threshold.

Whereas the locomotion rewards are concerned with the step timing of the robot, the gait-shaping rewards determine the spatial characteristics of the gait. 
The base pose reward is used to keep the roll and pitch of the base close to upright, 
and the base height close to a nominal desired base height. 
The stance foot flat term rewards the robot for keeping the foot pitch, $\boldsymbol{\Theta}_\textnormal{F,y}\in \mathbb{R}^{2}$, close to zero and is only active when the twist command is zero. 
For the hands, we reward the policy for tracking nominal hand trajectories, $\mathbf{p}_\textnormal{H}^\textnormal{d}$, that are a function of $v_\textnormal{x}^\textnormal{cmd}$. 
Although the hand positions have minimal effect on the SRB dynamics, as we do not model limb inertia,
we find that the simple arm motions aid in the Stage 2 training by providing references that are close to natural human arm swinging.

The remaining set of rewards are regularization rewards.
In Table. \ref{tab:rom_rewards} we define the base twist as $\boldsymbol{\xi} \triangleq \left[ \mathbf{v}_\textnormal{B} \ \boldsymbol{\omega}_\textnormal{B} \right] \in \mathbb{R}^6$ and $\boldsymbol{\zeta} \triangleq \left[ \boldsymbol{\xi} \ \dot{\boldsymbol{\xi}} \right] \in \mathbb{R}^{12}$.
For the normalized leg torque regularization we scale each torque value by the maximum torque of the respective joint. 
We use a power of $6$ to mainly penalize joint torques that are close to the max-joint torques.

\begin{figure*}
    \centering
    \includegraphics[width=1.0\linewidth]{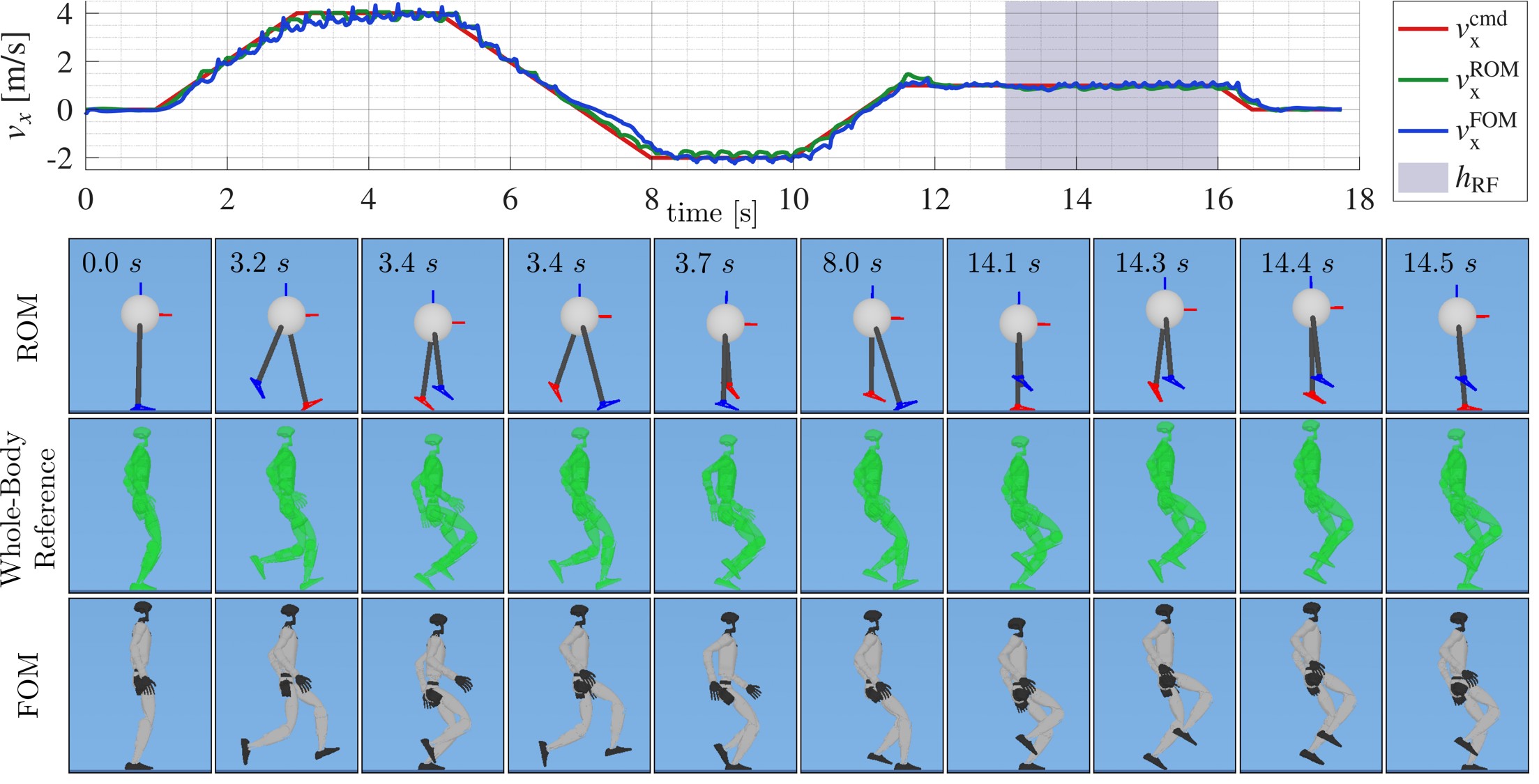}
    \caption{The experiment shows how the ROM and FOM track time-varying longitudinal twist ($v^\mathrm{cmd}_x$) and hopping commands ($h_\mathrm{RF}$).
    It also illustrates the ROM motions, whole-body reference, and tracked FOM motions on the simulated robot.}
    \label{fig:velocity_tracking}
    \vspace{-1.5em}
\end{figure*}

\section{Whole-Body Tracking}
\label{sec:method:stage2}

In the second stage, we train a whole-body
tracking policy that follows the command-conditioned reference produced online by the Stage-1 ROM-based reference generator. Our whole-body tracking frameworks builds on the BeyondMimic setup \cite{liao2025beyondmimic}. 

We use the following group of observations:
\begin{equation}
    \label{eq:fom_observation}
    \begin{aligned}
    \mathbf{o}_t^\textnormal{FOM} = \left[\mathbf{o}_{\mathrm{mimic}} \; \mathbf{o}_{\mathrm{phase}}^\textnormal{H} \;
    \mathbf{o}_{\mathrm{cmd}}^\textnormal{H}\right],
    \end{aligned}
\end{equation}
where $\mathbf{o}_{\mathrm{mimic}}$ is adopted from BeyondMimic unchanged, which include the whole-body kinematic reference, proprioception, and the previous action. We extend the observations with 
$\mathbf{o}_{\mathrm{phase}}$ and $\mathbf{o}_{\mathrm{cmd}}$ which are the phase and
command observations from \eqref{eq:rom_observation}, supplied by the online ROM planner.
The ROM observations are stacked over a history of $\textnormal{H} = 3$ control steps so that the policy observes the recent
evolution of the planner state rather than a single snapshot.

The reward terms are adopted from BeyondMimic~\cite{liao2025beyondmimic}. 
In addition, we augment the rewards with a twist-command tracking reward given by:
\begin{equation}
    r_{\mathrm{twist}} = w_\textnormal{twist}
    \sum_{i\in\{v_x,v_y,\omega_z\}}
    \exp\!(-(i-i^{\mathrm{cmd}})^2/\sigma^2_\textnormal{twist}),
\end{equation}
where $ w_\textnormal{twist} = 0.5$ and $\sigma_\textnormal{twist} = 0.2$.  
Although the ROM supplies a high-quality whole-body reference for the humanoid to imitate, 
a dynamics discrepancy between the reduced-order reference and the true full-body dynamics exists. 
This term helps the whole-body tracking policy in following the commanded twist. 
To reduce the joint torques, we also include the normalized leg torque penalty from the ROM training (Table~\ref{tab:rom_rewards}), evaluated using the FOM leg joint torques.





\section{Results}
\label{sec:results}

We evaluate the versatility and agility of the proposed two-stage framework, which couples online ROM-based reference generation with whole-body tracking control, through both simulation studies and hardware experiments on the Unitree G1 humanoid robot. Both the ROM reference generator and whole-body tracking controller operate at 50 Hz.

\subsection{Reference Generation and Tracking in Simulation}
\label{sec:results_sim}

In simulation, we verify that the ROM-guided reference generator produces whole-body
references that the Stage~2 policy can track while following the commanded velocity.
Particular emphasis is placed on continuously time-varying longitudinal commands and concurrent hopping
commands, both of which are handled by a single unified policy.
Figure~\ref{fig:velocity_tracking} depicts the commanded velocity ($v^\mathrm{cmd}_x$), the
velocity of the ROM reference ($v^\mathrm{ROM}_x$), and the velocity of the tracked
full-order model ($v^\mathrm{FOM}_x$) as the command ramps up to $4.0~\mathrm{m/s}$ for
fast running, reverses to $-2.0~\mathrm{m/s}$ for backward running, and settles at
$1.0~\mathrm{m/s}$ for walking before returning to rest.
The ROM reference follows the command closely over the full velocity envelope, confirming
that the Stage~1 generator produces feasible whole-body references across the entire command
range. The FOM in turn tracks the whole-body reference, so that the realized robot velocity
matches the command throughout these rapid transitions.
In the shaded interval, a right-foot air (hopping) command $h_\mathrm{RF}$ is issued
while the forward velocity command is held at $1.0~\mathrm{m/s}$; velocity tracking is
retained through this contact-schedule change, demonstrating that the framework composes
twist and hopping commands rather than treating them as separate modes.
The synchronized snapshots render this flow visible at both high speed and during hopping:
the ROM state drives the generation of a whole-body reference, which the Stage~2 policy
tracks to realize the corresponding robot motion.

\subsection{Hardware Experiments for Versatile Command Tracking}
\label{sec:results_tracking}

\begin{figure}
    \centering
    \includegraphics[width=\linewidth]{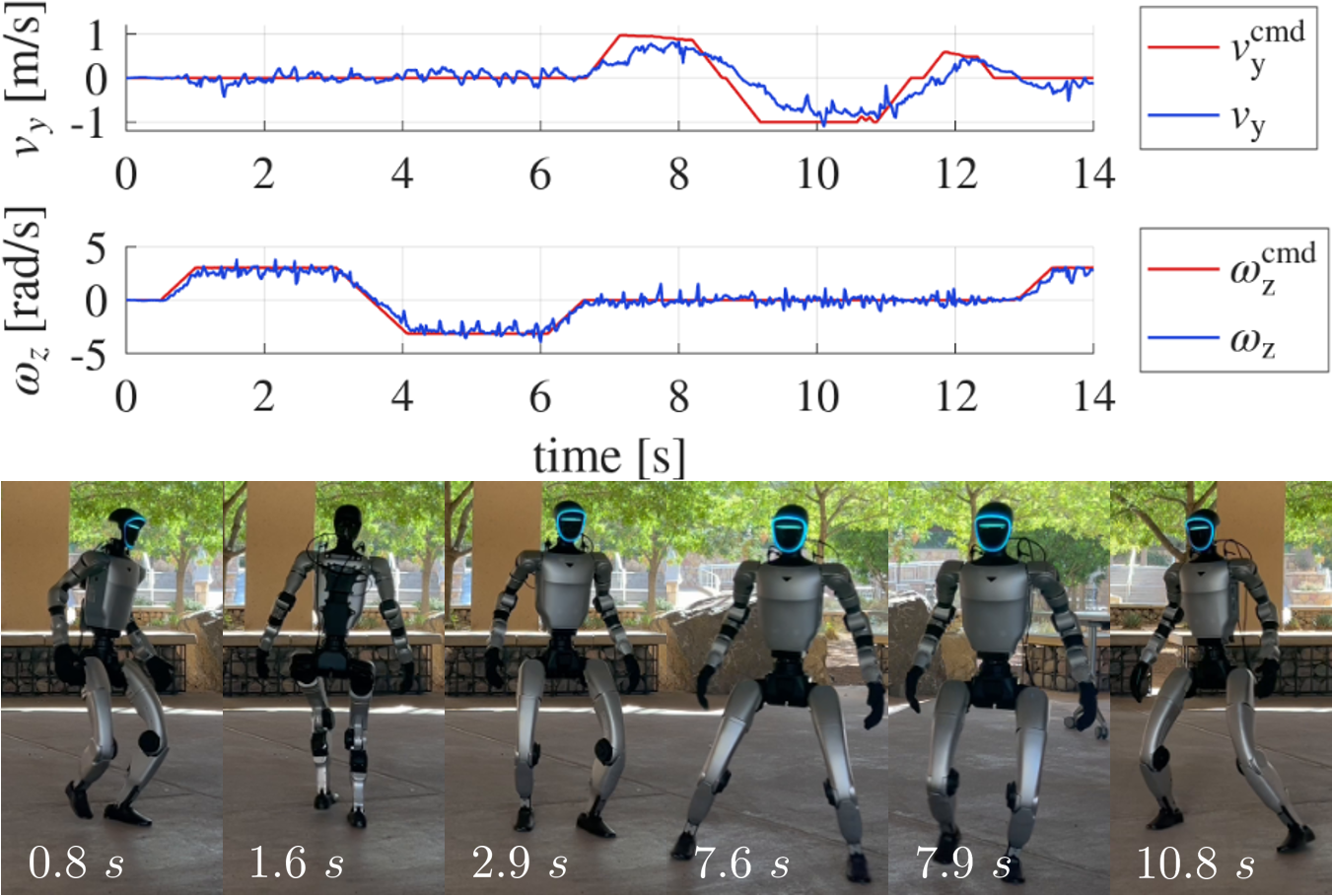}
    \caption{Hardware experiments tracking time varying lateral ($v_y$), and angular ($\omega_z$) velocity commands. Snapshots at $0.8$--$2.9~\mathrm{s}$ and at $7.6$--$10.8~\mathrm{s}$ show the corresponding turning and lateral stepping motions, respectively.
    }
    \label{fig:inputs}
\end{figure}
We demonstrate zero-shot sim-to-real transfer of our unified policy under time-varying
velocity commands. Figure~\ref{fig:inputs} shows the tracking performance for the lateral
($v_y^\mathrm{cmd}$) and angular ($\omega_z^\mathrm{cmd}$) velocity commands, which are varied
online, exercising yaw rates up to $\pm3.0~\mathrm{rad/s}$ and lateral velocities up to
$\pm1.0~\mathrm{m/s}$ on hardware. The velocities of Unitree G1 are measured from its onboard IMU and odometry estimator. The measured velocities follow the commanded signals
closely across both channels, and the robot transitions continuously between commands,
including sign reversals through zero, without switching between separately
trained skills. We emphasize that all results are produced by a single policy with no
per-task tuning, no gait-specific gain scheduling, and no mode-switching logic: lateral
stepping and in-place turning emerge from the same ROM-guided reference generator that
produces forward and backward walking. These results indicate that the whole-body references
supplied by the Stage~1 generator remain consistently trackable by the Stage~2 policy
across a broad command space, and that this property is preserved under transfer to hardware without additional tuning.

\subsection{Diverse Real-World Locomotion}
\label{sec:results_demos}

\begin{figure}
    \centering
    \includegraphics[width=1.0\linewidth]{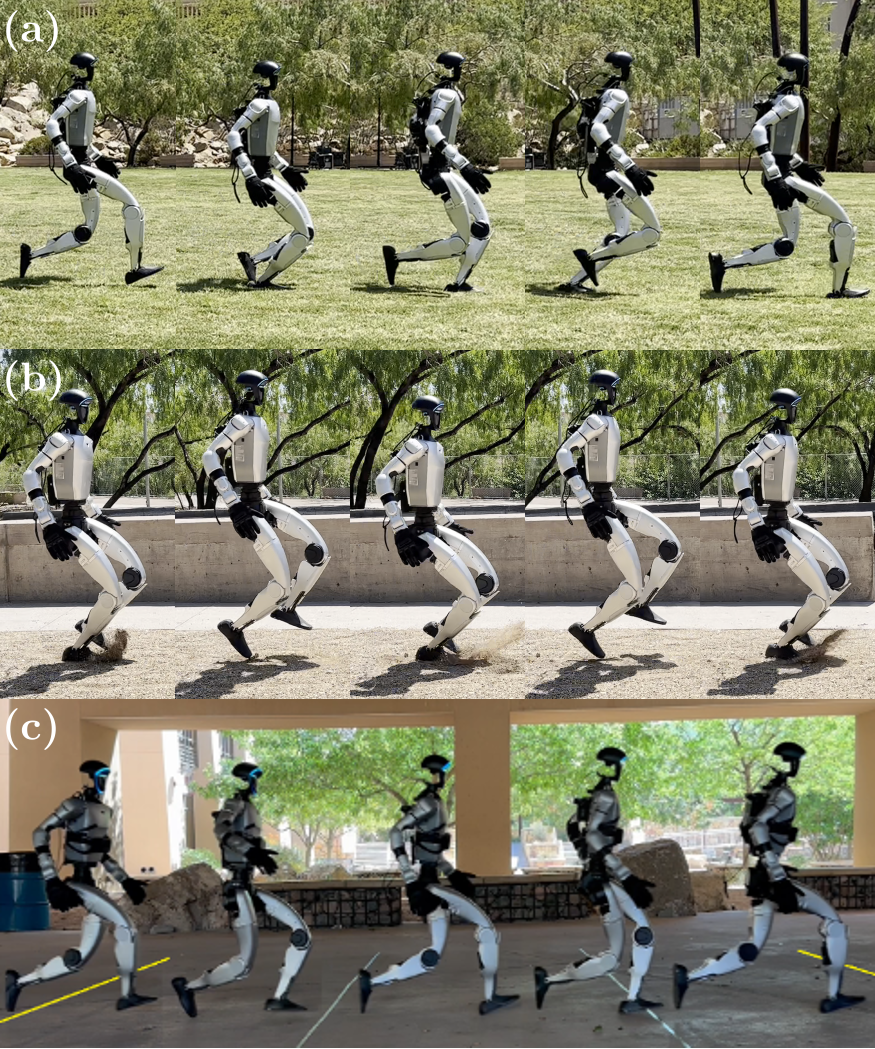}
    \caption{Snapshots of hardware demonstrations: 
    (a) Running on sloped uneven terrain.
    (b) One legged hopping on sand. 
    (c) Running at $2.8~\mathrm{m/s}$.}
    \label{fig:demo}
\end{figure}

To further evaluate robustness and dynamic behaviors of our unified policy, we conducted additional real-world
tests with the same unified policy. 
In Fig. \ref{fig:demo}(a) the robot performs high-speed running on uneven, sloped grassy terrain.
In Fig. \ref{fig:demo}(b) the robot is commanded to walk over sandy terrain, transitioned from walking to highly dynamic single-leg hopping, and subsequently resume walking. 
Although the policy was trained only on flat ground and had not encountered any of these terrain variations in training, the robot achieved stable locomotion in all cases. These results indicate that a
single policy trained within the proposed framework generalizes beyond the
nominal training conditions, handling diverse locomotion tasks in the real world. 
We also performed high-speed running experiments to validate the top speed achieved on hardware. 
Figure \ref{fig:demo}(c) shows an experiment where the robot is commanded to run at $v_\textnormal{x}^\textnormal{cmd} = 3.0~\mathrm{m/s}$. As the odometry data were noisy for high-velocity motions, we instead measured the average running speed over a distance of 15 feet (indicated by the yellow lines), and the robot was able to run at approximately $2.8~\mathrm{m/s}$.
%
It is noted that the results for fast running at $4.0 \mathrm{m/s}$ and sustained hopping are demonstrated only in simulation, as these motions approach the actuation capabilities of the physical robot and are therefore beyond the scope of the current hardware validation.


\section{Conclusion}

In this work we presented a framework for realizing versatile and agile locomotion on humanoid robots. 
We trained a command-conditioned whole-body reference generator based on ROM control, 
where SRB dynamics were combined with the full humanoid kinematics to synthesize the reference motions. 
Integrating this reference generator into a whole-body tracking framework 
enabled a single policy to realize a broad range of twist commands and multiple modes of locomotion. 
We demonstrated the proposed approach on a Unitree G1 humanoid robot, 
achieving longitudinal speeds from $-2.0$ m/s to $2.8$ m/s, 
lateral speeds up to $\pm 1.0$ m/s, turning rates up to $\pm 3.0$ rad/s, 
and the ability to hop on one leg. 
Future work will include ablation studies, comparisons with baseline controllers, and tracking performance analysis. 


\balance
\bibliographystyle{IEEEtran}
\newpage
\bibliography{99_references}

@inproceedings{reher2016realizing,
  title={Realizing dynamic and efficient bipedal locomotion on the humanoid robot DURUS},
  author={Reher, Jacob and Cousineau, Eric A and Hereid, Ayonga and Hubicki, Christian M and Ames, Aaron D},
  booktitle={2016 IEEE International Conference on Robotics and Automation (ICRA)},
  pages={1794--1801},
  year={2016},
  organization={IEEE}
}

@inproceedings{ma2017bipedal,
  title={Bipedal robotic running with {DURUS-2D}: Bridging the gap between theory and experiment},
  author={Ma, Wen-Loong and Kolathaya, Shishir and Ambrose, Eric R and Hubicki, Christian M and Ames, Aaron D},
  booktitle={Proc. Int. Conf. Hybrid Syst.: Comput. Control (HSCC)},
  pages={265--274},
  year={2017}
}

@article{ramezani2014performance,
  title={Performance analysis and feedback control of ATRIAS, a three-dimensional bipedal robot},
  author={Ramezani, Alireza and Hurst, Jonathan W and Akbari Hamed, Kaveh and Grizzle, Jessy W},
  journal={Journal of Dynamic Systems, Measurement, and Control},
  volume={136},
  number={2},
  pages={021012},
  year={2014},
  publisher={American Society of Mechanical Engineers}
}

@inproceedings{esteban2025reduced,
  title={Reduced-order model guided contact-implicit model predictive control for humanoid locomotion},
  author={Esteban, Sergio A and Kurtz, Vince and Ghansah, Adrian B and Ames, Aaron D},
  booktitle={2025 IEEE International Conference on Robotics and Automation (ICRA)},
  pages={14735--14741},
  year={2025},
  organization={IEEE}
}

@inproceedings{li2021force,
  title={Force-and-moment-based model predictive control for achieving highly dynamic locomotion on bipedal robots},
  author={Li, Junheng and Nguyen, Quan},
  booktitle={2021 60th IEEE Conference on Decision and Control (CDC)},
  pages={1024--1030},
  year={2021},
  organization={IEEE}
}

@inproceedings{ghansah2025hierarchical,
  title={Hierarchical Reduced-Order Model Predictive Control for Robust Locomotion on Humanoid Robots},
  author={Ghansah, Adrian B and Esteban, Sergio A and Ames, Aaron D},
  booktitle={2025 IEEE-RAS 24th International Conference on Humanoid Robots (Humanoids)},
  pages={1--8},
  year={2025},
  organization={IEEE}
}

@article{xiong20223,
  title={3-d underactuated bipedal walking via h-lip based gait synthesis and stepping stabilization},
  author={Xiong, Xiaobin and Ames, Aaron},
  journal={IEEE Transactions on Robotics},
  volume={38},
  number={4},
  pages={2405--2425},
  year={2022},
  publisher={IEEE}
}

@inproceedings{xiong2021global,
  title={Global position control on underactuated bipedal robots: Step-to-step dynamics approximation for step planning},
  author={Xiong, Xiaobin and Reher, Jenna and Ames, Aaron D},
  booktitle={2021 IEEE International Conference on Robotics and Automation (ICRA)},
  pages={2825--2831},
  year={2021},
  organization={IEEE}
}

@article{radosavovic2024real,
  title={Real-world humanoid locomotion with reinforcement learning},
  author={Radosavovic, Ilija and Xiao, Tete and Zhang, Bike and Darrell, Trevor and Malik, Jitendra and Sreenath, Koushil},
  journal={Science Robotics},
  volume={9},
  number={89},
  pages={eadi9579},
  year={2024},
  publisher={American Association for the Advancement of Science}
}

@inproceedings{siekmann2021sim,
  title={Sim-to-real learning of all common bipedal gaits via periodic reward composition},
  author={Siekmann, Jonah and Godse, Yesh and Fern, Alan and Hurst, Jonathan},
  booktitle={Proc. IEEE Int. Conf. Robot. Autom. (ICRA)},
  pages={7309--7315},
  year={2021}
}

@article{li2025reinforcement,
  title={Reinforcement learning for versatile, dynamic, and robust bipedal locomotion control},
  author={Li, Zhongyu and Peng, Xue Bin and Abbeel, Pieter and Levine, Sergey and Berseth, Glen and Sreenath, Koushil},
  journal={The International Journal of Robotics Research},
  volume={44},
  number={5},
  pages={840--888},
  year={2025},
  publisher={Sage Publications Sage UK: London, England}
}

@inproceedings{van2024revisiting,
  title={Revisiting reward design and evaluation for robust humanoid standing and walking},
  author={Van Marum, Bart and Shrestha, Aayam and Duan, Helei and Dugar, Pranay and Dao, Jeremy and Fern, Alan},
  booktitle={2024 IEEE/RSJ International Conference on Intelligent Robots and Systems (IROS)},
  pages={11256--11263},
  year={2024},
  organization={IEEE}
}

@article{liao2025beyondmimic,
  title={Beyondmimic: From motion tracking to versatile humanoid control via guided diffusion},
  author={Liao, Qiayuan and Truong, Takara E and Huang, Xiaoyu and Gao, Yuman and Tevet, Guy and Sreenath, Koushil and Liu, C Karen},
  journal={arXiv preprint arXiv:2508.08241},
  year={2025}
}

@article{sleiman2026zest,
  title={ZEST: Zero-shot Embodied Skill Transfer for Athletic Robot Control},
  author={Sleiman, Jean Pierre and Li, He and Adu-Bredu, Alphonsus and Deits, Robin and Kumar, Arun and Bergamin, Kevin and Bhardwaj, Mohak and Biddlestone, Scott and Burger, Nicola and Estrada, Matthew A and others},
  journal={arXiv preprint arXiv:2602.00401},
  year={2026}
}

@article{he2025asap,
  title={Asap: Aligning simulation and real-world physics for learning agile humanoid whole-body skills},
  author={He, Tairan and Gao, Jiawei and Xiao, Wenli and Zhang, Yuanhang and Wang, Zi and Wang, Jiashun and Luo, Zhengyi and He, Guanqi and Sobanbab, Nikhil and Pan, Chaoyi and others},
  journal={arXiv preprint arXiv:2502.01143},
  year={2025}
}

@article{wang2026spark,
  title={SPARK: Skeleton-Parameter Aligned Retargeting on Humanoid Robots with Kinodynamic Trajectory Optimization},
  author={Wang, Hanwen and Liao, Qiayuan and Zhang, Bike and Ren, Kunzhao and Sreenath, Koushil and Xiong, Xiaobin},
  journal={arXiv preprint arXiv:2603.11480},
  year={2026}
}

@inproceedings{kajita2003biped,
  title={Biped walking pattern generation by using preview control of zero-moment point},
  author={Kajita, Shuuji and Kanehiro, Fumio and Kaneko, Kenji and Fujiwara, Kiyoshi and Harada, Kensuke and Yokoi, Kazuhito and Hirukawa, Hirohisa},
  booktitle={2003 IEEE international conference on robotics and automation (Cat. No. 03CH37422)},
  volume={2},
  pages={1620--1626},
  year={2003},
  organization={IEEE}
}

@inproceedings{kajita20013d,
  title={The 3D linear inverted pendulum mode: A simple modeling for a biped walking pattern generation},
  author={Kajita, Shuuji and Kanehiro, Fumio and Kaneko, Kenji and Yokoi, Kazuhito and Hirukawa, Hirohisa},
  booktitle={Proceedings 2001 IEEE/RSJ International Conference on Intelligent Robots and Systems. Expanding the Societal Role of Robotics in the the Next Millennium (Cat. No. 01CH37180)},
  volume={1},
  pages={239--246},
  year={2001},
  organization={IEEE}
}

@article{gong2020angular,
  title={Angular momentum about the contact point for control of bipedal locomotion: Validation in a lip-based controller},
  author={Gong, Yukai and Grizzle, Jessy},
  journal={arXiv preprint arXiv:2008.10763},
  year={2020}
}

@inproceedings{dosunmu2023stair,
  title={Stair climbing using the angular momentum linear inverted pendulum model and model predictive control},
  author={Dosunmu-Ogunbi, Oluwami and Shrivastava, Aayushi and Gibson, Grant and Grizzle, Jessy W},
  booktitle={Proc. IEEE/RSJ Int. Conf. Intell. Robots Syst. (IROS)},
  pages={8558--8565},
  year={2023}
}

@article{blickhan1989spring,
  title={The spring-mass model for running and hopping},
  author={Blickhan, Reinhard},
  journal={Journal of biomechanics},
  volume={22},
  number={11-12},
  pages={1217--1227},
  year={1989},
  publisher={Elsevier}
}

@inproceedings{di2018dynamic,
  title={Dynamic locomotion in the mit cheetah 3 through convex model-predictive control},
  author={Di Carlo, Jared and Wensing, Patrick M and Katz, Benjamin and Bledt, Gerardo and Kim, Sangbae},
  booktitle={2018 IEEE/RSJ international conference on intelligent robots and systems (IROS)},
  pages={1--9},
  year={2018},
  organization={IEEE}
}

@book{raibert1986legged,
  title={Legged robots that balance},
  author={Raibert, Marc H},
  year={1986},
  publisher={MIT press}
}

@inproceedings{lee2024integrating,
  title={Integrating model-based footstep planning with model-free reinforcement learning for dynamic legged locomotion},
  author={Lee, Ho Jae and Hong, Seungwoo and Kim, Sangbae},
  booktitle={2024 IEEE/RSJ International Conference on Intelligent Robots and Systems (IROS)},
  pages={11248--11255},
  year={2024},
  organization={IEEE}
}

@article{liu2026hlip,
  title={HLIP-Guided Reinforcement Learning for Agile Multigait Biped Locomotion},
  author={Liu, Yan and Zang, Xizhe and Zhang, Xuehe and Song, Chao and Chen, Boyang and Zhang, Yue and Zhao, Jie},
  journal={IEEE Transactions on Industrial Electronics},
  year={2026},
  publisher={IEEE}
}

@article{dai2026walk,
  title={Walk the PLANC: Physics-Guided RL for Agile Humanoid Locomotion on Constrained Footholds},
  author={Dai, Min and Compton, William D and Li, Junheng and Yang, Lizhi and Ames, Aaron D},
  journal={arXiv preprint arXiv:2601.06286},
  year={2026}
}

@inproceedings{kim2025learning,
  title={Learning a Vision-Based Footstep Planner for Hierarchical Walking Control},
  author={Kim, Minku and Acosta, Brian and Chaudhari, Pratik and Posa, Michael},
  booktitle={2025 IEEE-RAS 24th International Conference on Humanoid Robots (Humanoids)},
  pages={1--8},
  year={2025},
  organization={IEEE}
}

@article{green2021learning,
  title={Learning spring mass locomotion: Guiding policies with a reduced-order model},
  author={Green, Kevin and Godse, Yesh and Dao, Jeremy and Hatton, Ross L and Fern, Alan and Hurst, Jonathan},
  journal={IEEE Robotics and Automation Letters},
  volume={6},
  number={2},
  pages={3926--3932},
  year={2021},
  publisher={IEEE}
}

@inproceedings{batke2022optimizing,
  title={Optimizing bipedal maneuvers of single rigid-body models for reinforcement learning},
  author={Batke, Ryan and Yu, Fangzhou and Dao, Jeremy and Hurst, Jonathan and Hatton, Ross L and Fern, Alan and Green, Kevin},
  booktitle={Proc. IEEE-RAS Int. Conf. Humanoid Robots (Humanoids)},
  pages={714--721},
  year={2022}
}

@article{compton2026terrain,
  title={Terrain Consistent Reference-Guided RL for Humanoid Navigation Autonomy},
  author={Compton, William D and Olkin, Zachary and Ames, Aaron D},
  journal={arXiv preprint arXiv:2605.15517},
  year={2026}
}

@inproceedings{todorov2012mujoco,
  title={Mujoco: A physics engine for model-based control},
  author={Todorov, Emanuel and Erez, Tom and Tassa, Yuval},
  booktitle={2012 IEEE/RSJ international conference on intelligent robots and systems},
  pages={5026--5033},
  year={2012},
  organization={IEEE}
}

@article{zhang2026wholebody,
  title={Learning Whole-Body Humanoid Locomotion via Motion Generation and Motion Tracking},
  author={Zhang, Zewei and Wen, Kehan and Xu, Michael and He, Junzhe and Li, Chenhao and Miki, Takahiro and Schwarke, Clemens and Zhang, Chong and Peng, Xue Bin and Hutter, Marco},
  journal={IEEE Robotics and Automation Letters},
  year={2026},
  publisher={IEEE}
}

@article{peng2018deepmimic,
  title={DeepMimic: Example-Guided Deep Reinforcement Learning of Physics-Based Character Skills},
  author={Peng, Xue Bin and Abbeel, Pieter and Levine, Sergey and van de Panne, Michiel},
  journal={ACM Transactions on Graphics},
  volume={37},
  number={4},
  pages={143:1--143:14},
  year={2018},
  publisher={ACM}
}

@inproceedings{chentanez2018physics,
  title={Physics-based Motion Capture Imitation with Deep Reinforcement Learning},
  author={Chentanez, Nuttapong and M{\"u}ller, Matthias and Macklin, Miles and Makoviychuk, Viktor and Jeschke, Stefan},
  booktitle={Proceedings of the 11th ACM SIGGRAPH Conference on Motion, Interaction and Games (MIG)},
  year={2018},
  publisher={ACM}
}

@article{yang2025omniretarget,
  title={OmniRetarget: Interaction-Preserving Data Generation for Humanoid Whole-Body Loco-Manipulation and Scene Interaction},
  author={Yang, Lujie and Huang, Xiaoyu and Wu, Zhen and Kanazawa, Angjoo and Abbeel, Pieter and Sferrazza, Carmelo and Liu, C. Karen and Duan, Rocky and Shi, Guanya},
  journal={arXiv preprint arXiv:2509.26633},
  year={2025}
}

@article{ze2025twist,
  title={TWIST: Teleoperated Whole-Body Imitation System},
  author={Ze, Yanjie and Chen, Zixuan and Ara{\'u}jo, Jo{\~a}o Pedro and Cao, Zi-ang and Peng, Xue Bin and Wu, Jiajun and Liu, C. Karen},
  journal={arXiv preprint arXiv:2505.02833},
  year={2025}
}

@article{ze2025twist2,
  title={TWIST2: Scalable, Portable, and Holistic Humanoid Data Collection System},
  author={Ze, Yanjie and Zhao, Siheng and Wang, Weizhuo and Kanazawa, Angjoo and Duan, Rocky and Abbeel, Pieter and Shi, Guanya and Wu, Jiajun and Liu, C. Karen},
  journal={arXiv preprint arXiv:2511.02832},
  year={2025}
}

@article{he2024h2o,
  title={Learning Human-to-Humanoid Real-Time Whole-Body Teleoperation},
  author={He, Tairan and Luo, Zhengyi and Xiao, Wenli and Zhang, Chong and Kitani, Kris and Liu, Changliu and Shi, Guanya},
  journal={arXiv preprint arXiv:2403.04436},
  year={2024}
}

@inproceedings{xu2025parc,
  title={PARC: Physics-based Augmentation with Reinforcement Learning for Character Controllers},
  author={Xu, Michael and Shi, Yi and Yin, KangKang and Peng, Xue Bin},
  booktitle={ACM SIGGRAPH 2025 Conference Papers},
  year={2025},
  publisher={ACM},
  doi={10.1145/3721238.3730616}
}

@inproceedings{serifi2024robotmdm,
  title={Robot Motion Diffusion Model: Motion Generation for Robotic Characters},
  author={Serifi, Agon and Grandia, Ruben and Knoop, Espen and Gross, Markus and B{\"a}cher, Moritz},
  booktitle={SIGGRAPH Asia 2024 Conference Papers},
  year={2024},
  publisher={ACM}
}

@article{liu2024opt2skill,
  title={Opt2Skill: Imitating Dynamically-feasible Whole-Body Trajectories for Versatile Humanoid Loco-Manipulation},
  author={Liu, Fukang and Gu, Zhaoyuan and Cai, Yilin and Zhou, Ziyi and Jung, Hyunyoung and Jang, Jaehwi and Zhao, Shijie and Ha, Sehoon and Chen, Yue and Xu, Danfei and Zhao, Ye},
  journal={IEEE Robotics and Automation Letters},
  year={2025},
  publisher={IEEE}
}


\end{document}